\documentclass[sigconf, nonacm]{acmart}
\usepackage{enumitem} 
\usepackage{lipsum}
\newcommand \footnoteONLYtext[1]
{
	\let \mybackup \thefootnote
	\let \thefootnote \relax
	\footnotetext{#1}
	\let \thefootnote \mybackup
	\let \mybackup \imareallyundefinedcommand
}

\begin{document}
\title{Inter- and Intra-Series Embeddings Fusion Network for Epidemiological Forecasting}

\author{Feng Xie}
\affiliation{%
  \institution{National University of Defense Technology}
  \streetaddress{College of Computer, National University of Defense Technology}
  \city{Changsha}
  \state{China}
  \postcode{410005}
}
\email{xiefeng@nudt.edu.cn}

\author{Zhong Zhang}
\affiliation{%
  \institution{National University of Defense Technology}
  \streetaddress{College of Computer, National University of Defense Technology}
  \city{Changsha}
  \state{China}
  \postcode{410005}
}
\email{zhangzhong@nudt.edu.cn}

\author{Xuechen Zhao}
\orcid{0000-0002-1825-0097}
\affiliation{%
  \institution{National University of Defense Technology}
  \streetaddress{College of Computer, National University of Defense Technology}
  \city{Changsha}
  \country{China}
}
\email{zhaoxuechen@nudt.edu.cn}

\author{Bin Zhou}
\authornote{Corresponding Author: binzhou@nudt.edu.cn}
\orcid{0000-0002-1825-0097}
\affiliation{%
  \institution{National University of Defense Technology}
  \streetaddress{College of Computer, National University of Defense Technology}
  \city{Changsha}
  \country{China}
}
\email{binzhou@nudt.edu.cn}

\author{Yusong Tan}
\orcid{0000-0002-1825-0097}
\affiliation{%
  \institution{National University of Defense Technology}
  \streetaddress{College of Computer, National University of Defense Technology}
  \city{Changsha}
  \country{China}
}
\email{ystan@nudt.edu.cn}

\begin{abstract}
The accurate forecasting of infectious epidemic diseases is the key to effective control of the epidemic situation in a region. Most existing methods ignore potential dynamic dependencies between regions or the importance of temporal dependencies and inter-dependencies between regions for prediction. In this paper, we propose an Inter- and Intra-\underline{\textbf{S}}eries \underline{\textbf{E}}mbeddings \underline{\textbf{F}}usion \underline{\textbf{Net}}work (SEFNet) to improve epidemic prediction performance. SEFNet consists of two parallel modules, named Inter-Series Embedding Module and Intra-Series Embedding Module. In Inter-Series Embedding Module, a multi-scale unified convolution component called Region-Aware Convolution is proposed, which cooperates with self-attention to capture dynamic dependencies between time series obtained from multiple regions. The Intra-Series Embedding Module uses Long Short-Term Memory to capture temporal relationships within each time series. Subsequently, we learn the influence degree of two embeddings and fuse them with the parametric-matrix fusion method. To further improve the robustness, SEFNet also integrates a traditional autoregressive component in parallel with nonlinear neural networks. Experiments on four real-world epidemic-related datasets show SEFNet is effective and outperforms state-of-the-art baselines.
\end{abstract}

\maketitle
\section*{KEYWORDS}
deep learning; epidemiological forecasting; time series

\section{Introduction}
The outbreak of an epidemic will bring huge disasters to a region and even a country. The World Health Organization (WHO) estimates that influenza annually causes approximately 3-5 million severe cases and 290,000-650,000 deaths.\footnote{https://www.who.int/en/news-room/fact-sheets/detail/influenza-(seasonal)} In recent years, the COVID-19 pandemic has spread to more than 200 countries and territories around the world,\footnote{https://covid19.who.int/} and the number of infections and deaths in almost all affected countries is increasing at an alarming rate. Accurately forecasting epidemics plays an essential role in allocating healthcare resources and promoting administrative planning.

\footnoteONLYtext{DOI reference number: 10.18293/SEKE2022-109}

The epidemic prediction is similar to the multivariate time series forecasting task, but there are also significant differences. Multivariate time series forecasting methods inherently assume inter-dependencies among variables ~\cite{wu2020connecting}, while epidemic prediction needs to deal with unknown and complex patterns in the spread of epidemics and dynamic correlations between regions. The epidemic situation of a region at a certain time step is correlated with both its previous confirmed cases and other regions' epidemic situation. Therefore, two types of dependencies can be utilized in time series as demonstrated by \autoref{fig:example} and the epidemic time series modeling of regions can be decomposed into two parts:


\begin{figure}[h]
  \centering
  \includegraphics[width=0.8\linewidth]{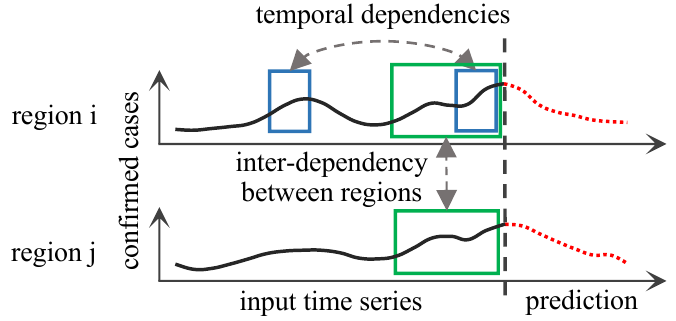}
  \caption{The blue boxes indicate the temporal dependency between time points, while the green boxes indicate the inter-dependency between regions.}
  \label{fig:example}
\end{figure}

\begin{figure*}[t]
	\centering 
	\includegraphics[width=0.78\linewidth]{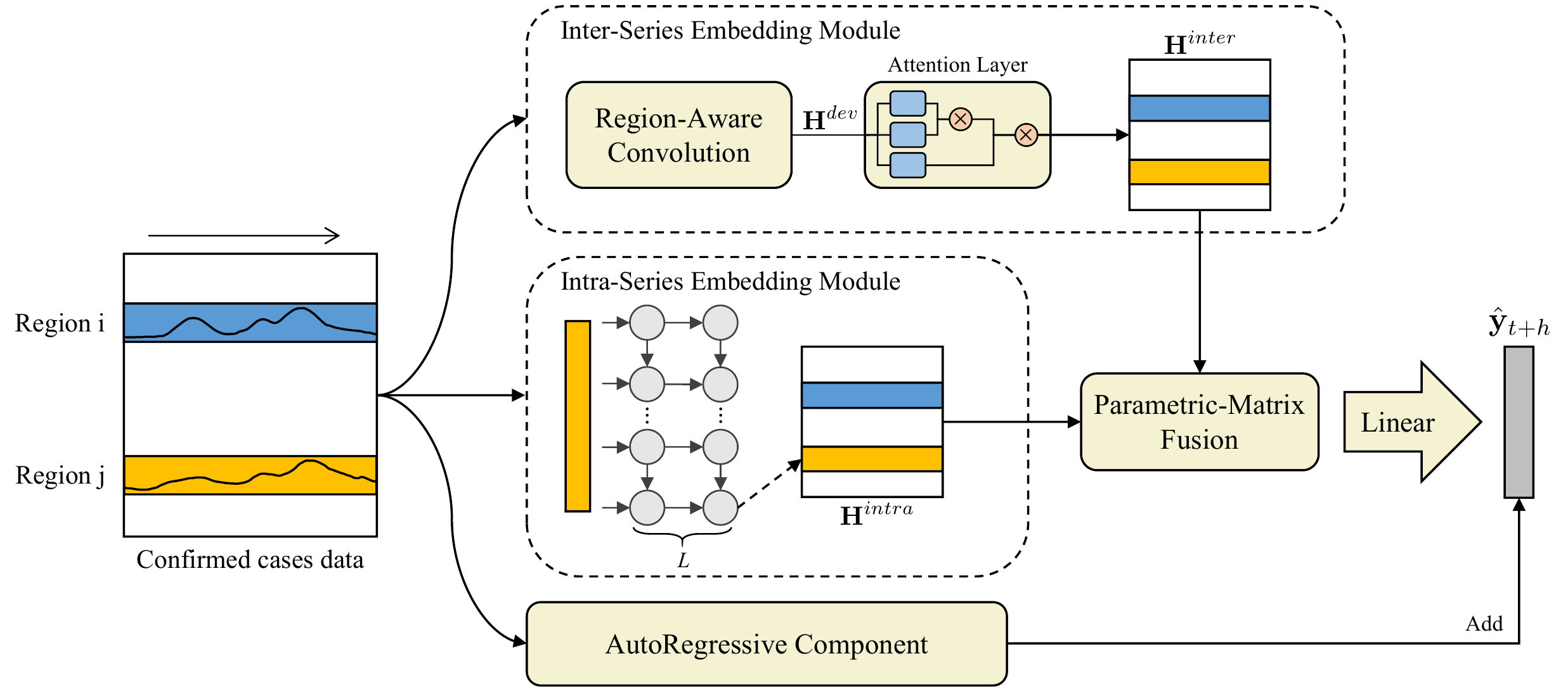}
	\caption{The overview of SEFNet. The original time series for each region are copied to three components: (1) Inter-Series Embedding module (top); (2) Intra-Series Embedding module (middle); and (3) AutoRegressive component (bottom).} 
	\label{fig:model}
\end{figure*}

\begin{itemize}[leftmargin=*]
    \item \textbf{Inter-series embedding modeling.} There are dynamic dependencies among regions/time series. When epidemics spread in different geographic regions, it is highly likely that similar progression patterns are shared among multiple regions owing to various factors \cite{jin2021inter} (e.g., similar geographic topology or climate), and these similar patterns can aid in prediction.
    \item \textbf{Intra-series embedding modeling.} There are temporal dependencies within a region/time series (e.g., seasonal influenza). The epidemic development trend of one region also can be distinguished from others, this is due to region-specific factors such as government intervention, healthcare quality, climate, etc.
\end{itemize}

To date, various methods have been proposed for epidemic forecasting, but they suffer from some limitations that are bad for performance. First, using vanilla Recurrent Neural Networks (RNNs) ~\cite{deng2020cola,jung2021self} or single-scale Convolution Neural Networks (CNNs) ~\cite{wu2018deep,jin2021inter} is hard to capture multi-scale and complex patterns, thus resulting in a certain degree of distortion, making it difficult to extract dynamic dependencies between time series. Second, some methods are dedicated to capturing dependencies between regions by introducing the "Attention Mechanism" ~\cite{jung2021self,jin2021inter,deng2020cola}, but these dependencies may misguide the final prediction because the progression patterns or data distribution of different regions is not fully consistent. Therefore, we believe both inter-series dependencies and intra-series dependencies jointly contribute to epidemic forecasting, and their influence degree on prediction results varies by region. 

To tackle these challenges, we propose a novel deep learning model called Inter- and Intra-\underline{\textbf{S}}eries \underline{\textbf{E}}mbeddings \underline{\textbf{F}}usion \underline{\textbf{Net}}work (SEFNet) that extract inter- and intra-series embeddings through two parallel modules respectively and fuse them using parametric-matrix fusion ~\cite{zhang2017deep}. To further improve the robustness, we also integrate autoregressive component parallel to the model. Our contributions are summarized below:

\begin{itemize}[leftmargin=*]
    \item We propose a new model that extracts inter-series correlations and intra-series temporal dependencies through two separate neural networks and uses parametric-matrix fusion to emphasize the importance of each information for epidemic prediction.
    \item We propose a multi-scale unified convolution component called Region-Aware Convolution that is capable of extracting local, periodic, and global patterns to better obtain feature representation and capture potential dependencies between regions.
    \item We conduct extensive experiments on four real-world epidemic-related datasets. The results show that our model achieves better performance than other state-of-the-art methods and demonstrates the effectiveness of each component.
\end{itemize}

\section{Related Work}
There has been a large body of work focusing on epidemic forecasting in literature, including statistical models \cite{martinez2011sarima,wang2015dynamic,jia2020population}, compartment models \cite{harko2014exact,won2017early}, and swarm intelligence models \cite{guo2021improving}. In recent years, deep learning models have shown excellent performance in various prediction tasks due to their powerful training and data-driven capabilities. CNNRNN-Res ~\cite{wu2018deep} is the first to apply deep learning for epidemiological prediction. ~\citet{deng2020cola} proposed Cola-GNN that treats regions as nodes in a graph and applies Graph Neural Networks (GNNs) to capture dependencies among regions. \citet{jung2021self} proposed SAIFlu-Net that combines Long Short-Term Memory and self-attention to capture inter-dependencies between regions. \citet{jin2021inter} developed ACTs based on inter-series attention for COVID-19 forecasting. \citet{cui2021into} designed a multi-range encoder-decoder framework for COVID-19 prediction. Nowadays, improving epidemic prediction is an open research problem to help the world mitigate the crisis that threatens public health.

\section{The Proposed Method}

\subsection{Problem Formulation}
We formulate the epidemic prediction problem as a time series forecasting task. We have a total of $N$ regions, and each region is associated with a time series input for a window $T$, where $T$ is the length of historical observation data. Furthermore, we denote the epidemiology profiles $\mathbf{X}=[\mathbf{x}_{t-T+1},...,\mathbf{x}_t]\in\mathbb{R}^{N\times{T}}$ at time point $t$. An instance for region $i$ is represented by $\mathbf{x}_{i:}=[x_{i,t-T+1},...,x_{i,t}]\in\mathbb{R}^{T}$. The goal of this task is to predict the epidemiology profile of the future time point $t+h$, where $h$ is the horizon also called lead time. The proposed model SEFNet is shown in \autoref{fig:model}. In the following sections, we introduce the building blocks of SEFNet in detail.

\subsection{Intra-Series Embedding Module}
The first module is Intra-Series Embedding module, which uses the historical information of time series to focus on the autocorrelation also called temporal dependencies of a single time series. In this work, we apply the Long Short-Term Memory (LSTM) ~\cite{hochreiter1997long} to capture temporal sequential dependency. The Recurrent Neural Networks (RNNs) are shown effective in sequence modeling and LSTM is a variant of RNNs, which can solve the vanishing gradients and exploding gradients problems in traditional RNNs \cite{siami2019performance}. Let $D$ be the dimension of the hidden state of LSTM, we use the original version of LSTM and formulate it as:
\begin{equation}
  \mathbf{h}_{i,t}=\text{LSTM}(x_{i,t},\mathbf{h}_{i,t-1}),
\end{equation}

\noindent where $\mathbf{h}_{i,t}$ is the output representation of region $i$ at time point $t$. For each region, we use the last output of LSTM as region's intra-series embedding $\mathbf{h}_{i}^{intra}\in\mathbb{R}^{D}$.




\subsection{Inter-Series Embedding Module}
The second module is Inter-Series Embedding module which focuses on dependencies between time series. First, we obtain temporal patterns through the proposed Region-Aware Convolution (RAConv), which is a multi-scale unified convolution component. Next, we feed the output of RAConv into an attention layer to generate embeddings of dynamic dependencies between regions. 

\begin{figure}[h]
  \centering
  \includegraphics[width=\linewidth]{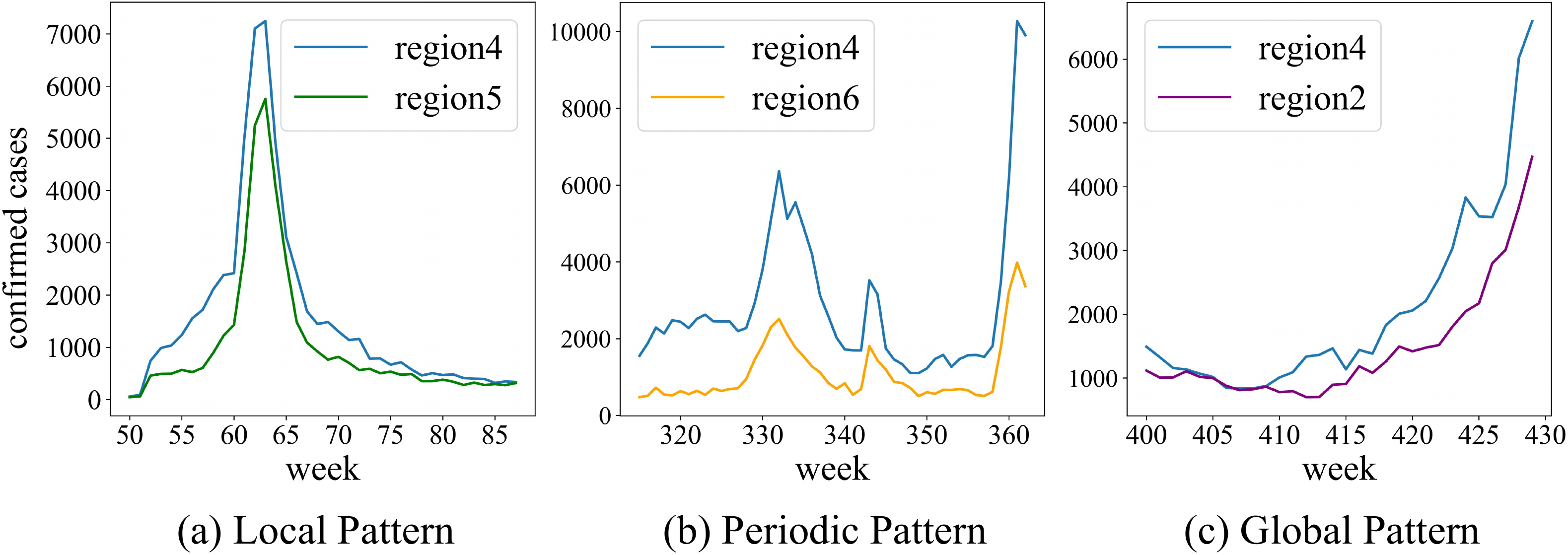}
  \caption{The region 4 (blue line) has dynamic pattern correlations with different regions within different time periods.}
  \label{fig:pattern}
\end{figure}

\begin{figure}
  \centering
  \includegraphics[width=\linewidth]{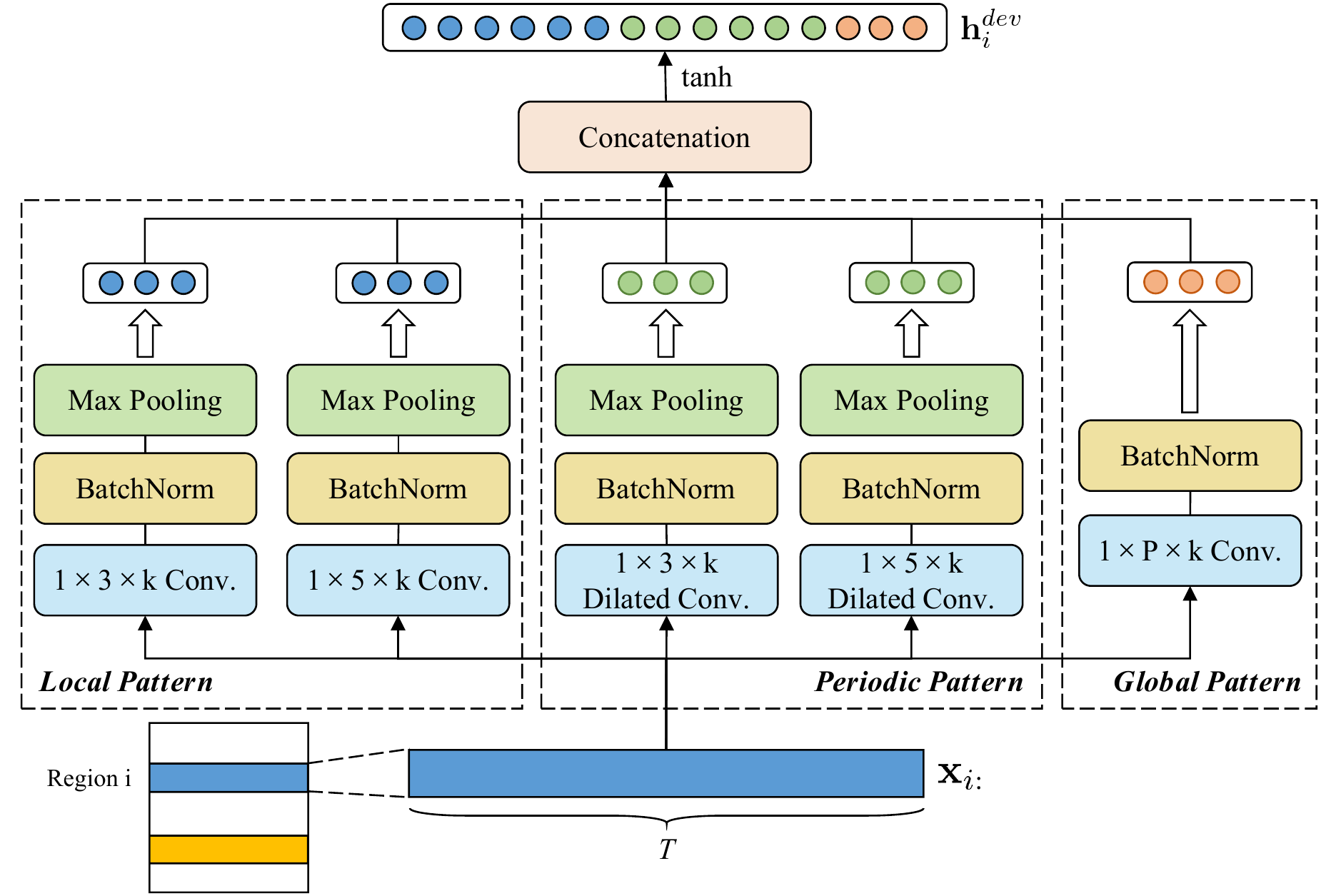}
  \caption{The structure of Region-Aware Convolution, which consists of three pattern branches.}
  \label{fig:rac}
\end{figure}

The correlation distribution is calculated based on the feature similarity between nodes \cite{DCRN}. The more accurate feature describes, the better performance of the attention layer can be improved. In epidemic prediction task, there are many similar progression patterns shared among regions, such as local patterns, periodic patterns, and global patterns. \autoref{fig:pattern} shows different temporal patterns of influenza case trends in different Health and Human Services (HHS) regions in the United States. Inspired by the Inception ~\cite{szegedy2015going} in computer vision, we propose a multi-scale unified component called Region-Aware Convolution (RAConv) that can extract local, periodic, and global patterns simultaneously. The structure of RAConv is shown in \autoref{fig:rac}. RAConv consists of three branches that apply convolution blocks with different scales or different types, thus is capable of capturing multi-scale and more complex feature patterns. Each convolution block has $K$ filters. The local pattern branch applies standard convolution with some small kernel sizes to extract local patterns in the time series through local mapping. The periodic pattern branch inspired by skip-RNN in LSTNet \cite{lai2018modeling} applies dilated convolution that enables a large receptive field via dilation factor to capture the periodic pattern. Formally, the dilated convolution is a standard convolution applied to input with defined gaps. The global pattern branch applies standard convolution with the same size as $T$ to extract time-invariant patterns of all time steps for regions \cite{huang2019dsanet} (e.g., time series uptrend in \autoref{fig:pattern}c). We denote convolution filter in RAConv as $f_{1\times{s},d}$ where $s$ is kernel size and $d$ is dilated factor. We empirically choose the kernel size $s$ to \{3,5,$T$\}, and dilated factor $d$ to \{1,2\}. We can get the local, periodic, and global features of region $i$ by following equations:
\begin{equation}
  \mathbf{h}_{i}^{l}=[\text{Pool}(\text{BN}(\mathbf{x}_{i:}\star{\mathbf{f}_{1\times{3},{1}}}));\text{Pool}(\text{BN}(\mathbf{x}_{i:}\star{\mathbf{f}_{1\times{5},{1}}}))],
\end{equation}
\begin{equation}
  \mathbf{h}_{i}^{p}=[\text{Pool}(\text{BN}(\mathbf{x}_{i:}\star{\mathbf{f}_{1\times{3},{2}}}));\text{Pool}(\text{BN}(\mathbf{x}_{i:}\star{\mathbf{f}_{1\times{5},{2}}}))],
\end{equation}
\begin{equation}
  \mathbf{h}_{i}^{g}=\text{BN}(\mathbf{x}_{i:}\star{\mathbf{f}_{1\times{T},{1}}}),
\end{equation}

\noindent where $\star$ is convolution operator, and $[;]$ is concatenation operation. $Pool(\cdot)$ is the Adaptive Max Pooling layer that can not only capture the most representative features, but also effectively reduce the amount of parameters. Adaptive Max Pooling is able to control the output size same as parameters $P$. $BN(\cdot)$ is the Batch Normalization ~\cite{ioffe2015batch} layer that normalize the data and speed up convergence. The convolution operation of $\mathbf{x}$ with $\mathbf{f}$ at step $j$ is represented as: 
\begin{equation}
  \mathbf{x}\star{\mathbf{f}_{1\times{s},{d}}}(j)=\sum_{i=1}^{s}\mathbf{f}_{1\times{k}}(i)\mathbf{x}(j-d\times{i}).
\end{equation}

Next, we concatenate three patterns and apply an element-wise activation function (e.g., hyperbolic tangent):
\begin{equation}
  \mathbf{h}_{i}^{dev}=\text{tanh}([\mathbf{h}_{i}^{l};\mathbf{h}_{i}^{p};\mathbf{h}_{i}^{g}]).
\end{equation}

For each time series, we execute the above process and get the intermediate matrix called $\mathbf{H}^{dev}\in\mathbb{R}^{N\times{(P+1)K}}$.

Due to the powerful feature extraction capability of the self-attention network, we apply a typical self-attention network inspired by the Transformer ~\cite{vaswani2017attention} to capture the dependencies among regions. Let $A$ be the dimension of inter-series embedding. We can calculate attention distribution $\mathbf{A}$ and inter-series embedding matrix $\mathbf{H}^{inter}\in\mathbb{R}^{N\times{A}}$ by following equations:
\begin{equation}
    \mathbf{A}=\text{softmax}((\mathbf{H}^{dev}\mathbf{W}^{Q})(\mathbf{H}^{dev}\mathbf{W}^{K})^T),
\end{equation}
\begin{equation}
    \mathbf{H}^{inter}=\mathbf{A}\mathbf{H}^{dev}\mathbf{W}^{V},
\end{equation}

\noindent where $\mathbf{W}^Q$, $\mathbf{W}^K$, and $\mathbf{W}^V \in\mathbb{R}^{(P+1)K\times{A}}$ are the weight matrices that linearly map the $\mathbf{H}^{dev}$ to query, key, and value matrices. 


\subsection{Fusion}
Directly concatenating or summing inter-series embedding and intra-series embedding will have the following problems: $(1)$ \textbf{Inconsistent scale.} Since two feature embeddings come from different neural network modules, the structural differences of each module (e.g., activation function) will lead to inconsistent scales of feature embeddings; $(2)$ \textbf{Different importance.} Two feature embeddings describe different feature information of time series so the importance of two feature embeddings is very different in the process of epidemiology forecasting (e.g, temporal dependency is more significant for a region with periodic recurrence of an epidemic, although there may be similar development patterns to others). Therefore, to address these problems, we adopt parametric-matrix fusion \cite{zhang2017deep} to adaptively control the flow of inter-series embedding and intra-series embedding and fuse them together:
\begin{equation}
  \mathbf{H}^{fus} = [\mathbf{W}^{inter}\circ \mathbf{H}^{inter} ; \mathbf{W}^{intra} \circ \mathbf{H}^{intra}],
\end{equation}

\noindent where $\mathbf{H}^{fus}\in\mathbb{R}^{N\times{(D+A)}}$ is the ouput of fusion operation. $\circ$ is element-wise multiplication. $\mathbf{W}^{inter}$ and $\mathbf{W}^{intra}$ are the learnable parameters that adjust the degrees affected by inter-series embedding and intra-series embedding respectively.

\subsection{Prediction}
Due to the nonlinear characteristics of Convolutional, Recurrent and self-attention components, the scale of neural network output is not sensitive to input. Meanwhile, the historical infection cases of each region are not purely nonlinear, which cannot be fully handled well by neural networks. To address these drawbacks, we retain the advantages of traditional linear models and neural networks by combining a linear part to design a more accurate and robust prediction framework inspired by ~\cite{lai2018modeling,shih2019temporal}. Specifically, we adopt the classical AutoRegressive (AR) model as the linear component in a parallel manner. Denote the forecasting result of AR component as $\mathbf{\hat{y}}^{l}_{t+h}\in\mathbb{R}^{N}$ that can be calculated by following equation:
\begin{equation}
  {\hat{y}}^{l}_{i,t+h} = \sum_{m=0}^{q-1}\mathbf{W}_m^{ar}x_{i,t-m}+b^{ar},
\end{equation}

\noindent where $\mathbf{W}^{ar}$ is the weight matrix and $b^{ar}$ is the bias. $q$ is the look-back window of AR that need be less than or equal to input window size $T$. Then, we feed the output after fusion operation to a dense layer to get the nonlinear part of the final prediction:
\begin{equation}
  \mathbf{\hat{y}}^{n}_{t+h} = \mathbf{H}^{fus}\mathbf{W}^{n} + \mathbf{b}^{n}.
\end{equation}

The final prediction of model is then obtained by summing the nonlinear part and the linear part got by AR component:
\begin{equation}
  \hat{\mathbf{y}}_{t+h} = \mathbf{\hat{y}}^{n}_{t+h} + \mathbf{\hat{y}}^{l}_{t+h}.
\end{equation}

In the training process, we adopt the Mean Square Error as the loss function that defined as:
\begin{equation}
  \underset{\theta}{minimize} \left\| \mathbf{y}_{t+h}-\hat{\mathbf{y}}_{t+h} \right\| ^2_2 ,
\end{equation}

\noindent where $\mathbf{y}_{t+h}=[y_{1,t+h},...,y_{N,t+h}]\in\mathbb{R}^{N}$ is the true value at time point $t+h$, and $\theta$ are all learnable parameters in the model. 

\begin{table}[h]
  \caption{Dataset statistics: min, max, mean, and standard deviation (SD) of patient counts; dataset size means the number of regions multiplied by the number of samples.}
  \label{tab:datasets}
  \begin{tabular}{lccccc}
    \toprule
    Datasets & Size & Min & Max & Mean & SD\\
    \midrule
    Japan-Prefectures & 47$\times$348 & 0 & 26635 & 655 & 1711 \\
    US-Regions & 10$\times$785 & 0 & 16526 & 1009 & 1351 \\
    US-States & 49$\times$360 & 0 & 9716 & 223 & 428 \\
    Canada-Covid & 13$\times$717 & 0 & 127199 & 3082 & 8473 \\
  \bottomrule
\end{tabular}
\end{table}

\section{Experiments}

\subsection{Datasets and Metrics}

We prepare four real-world epidemic-related datasets as follows, and their data statistics are shown in \autoref{tab:datasets}.

\begin{table*}[t]
  \small
  \caption{RMSE and PCC performance of different methods on four datasets with horizon = 3, 5, 10. Bold face indicates the best result of each column and underlined the second-best. For RMSE lower value is better, while for PCC higher value is better.}
  \label{tab:result}
  \renewcommand{\arraystretch}{0.8} 
  \setlength{\tabcolsep}{2.3mm}{
  \begin{tabular}{llc|ccc|ccc|ccc|ccc}
    \toprule
    Dataset & & & \multicolumn{3}{c|}{Japan-Prefectures} & \multicolumn{3}{c|}{US-Regions} & \multicolumn{3}{c|}{US-States} & \multicolumn{3}{c}{Canada-Covid}\\
    \midrule
    & & & \multicolumn{3}{c|}{Horizon} & \multicolumn{3}{c|}{Horizon} & \multicolumn{3}{c|}{Horizon} & \multicolumn{3}{c}{Horizon}\\
    \midrule
    Methods & Metrics & & 3 & 5 & 10 & 3 & 5 & 10 & 3 & 5 & 10 & 3 & 5 & 10\\
    \midrule
    AR & RMSE & & 1705 & 2013 & 2107 & 757 & 997 & 1330 & 204 & 251 & 306 & 3488 & 4545 & \underline{7154}\\
    & PCC & & 0.579 & 0.310 & 0.238 & 0.878 & 0.792 & 0.612 & 0.909 & 0.863 & 0.773 & 0.973 & 0.955 & 0.869\\
    \midrule
    LRidge & RMSE & & 1711 & 2025 & 1942 & 870 & 1059 & 1270 & 276 & 295 & 324 & 3326 & 4372 & 7179\\
    & PCC & & 0.308 & 0.429 & 0.238 & 0.878 & 0.792 & 0.612 & 0.909 & 0.863 & 0.773  & 0.975 & 0.957 & 0.868\\
    \midrule
    LSTNet & RMSE & & 1459 & 1883 & 1811 & 801 & 998 & 1157 & 249 & 299 & 292 & 3270 & 6789 & 9561\\
    & PCC & & 0.728 & 0.432 & 0.518 & 0.868 & 0.746 & 0.609 & 0.850 & 0.759 & 0.760 & 0.967 & 0.847 & 0.645\\
    \midrule
    TPA-LSTM & RMSE & & 1142 & 1192 & 1677 & 761 & 950 & 1388 & 203 & 247 & \underline{236} & \underline{2731} & \underline{3905} & 7671\\
    & PCC & & 0.879 & 0.868 & 0.644 & 0.847 & 0.814 & 0.675 & 0.892 & 0.833 & \underline{0.849} & 0.980 & 0.956 & 0.767\\
    \midrule
    CNNRNN-Res & RMSE & & 1550 & 1942 & 1865 & 738 & 936 & 1233 & 239 & 267 & 260  & 6175 & 8644 & 9755\\
    & PCC & & 0.673 & 0.380 & 0.438 & 0.862 & 0.782 & 0.552 & 0.860 & 0.822 & 0.820  & 0.659 & 0.589 & 0.475\\
    \midrule
    SAIFlu-Net & RMSE & & 1356 & 1430 & 1527 & 661 & 871 & 1158 & \underline{167} & 238 & \underline{236} & 4409 & 7128 & 8514\\
    & PCC & & 0.765 & 0.654 & 0.592 & \underline{0.903} & 0.800 & 0.674 & 0.927 & 0.842 & 0.845  & 0.745 & 0.775 & 0.596\\
    \midrule
    Cola-GNN & RMSE & & \underline{1051} & \textbf{1117} & \underline{1372} & \underline{636} & \underline{855} & \underline{1134} & \underline{167} & \underline{202} & 241  & 2954 & 4036 & 7336\\
    & PCC & & \underline{0.901} & \underline{0.890} & \underline{0.813} & \textbf{0.909} & \underline{0.835} & \underline{0.717} & \underline{0.933} & \underline{0.897} & 0.822 & \underline{0.986} & \underline{0.975} & \underline{0.882}\\
    \midrule
    \midrule
    SEFNet & RMSE & \footnotesize{($\downarrow$)} & \textbf{1020} & \underline{1123} & \textbf{1319} & \textbf{618} & \textbf{821} & \textbf{1036} & \textbf{162} & \textbf{196} & \textbf{232} & \textbf{2157} & \textbf{3339} & \textbf{7079}\\
    & PCC & \footnotesize{($\uparrow$)} & \textbf{0.904} & \textbf{0.893} & \textbf{0.826} & \textbf{0.909} & \textbf{0.842} & \textbf{0.725} & \textbf{0.935} & \textbf{0.900} & \textbf{0.833} & \textbf{0.990} & \textbf{0.978} & \textbf{0.895}\\
    \bottomrule
  \end{tabular}}
\end{table*}

\begin{itemize}[leftmargin=*]
\item \textbf{Japan-Prefectures.} This dataset is collected from the Infectious Diseases Weekly Report (IDWR) in Japan,\footnote{https://tinyurl.com/y5dt7stm} which contains weekly influenza-like-illness statistics from 47 prefectures, ranging from August 2012 to March 2019.
\item \textbf{US-Regions.} This dataset is the ILINet portion of the US-HHS dataset,$\footnote{https://tinyurl.com/y39tog3h\label{comm:us-data}}$ consisting of weekly influenza activity levels for 10 HHS regions of the U.S. mainland for the period of 2002 to 2017. Each HHS region represents some collection of associated states.
\item \textbf{US-States.} This dataset is collected from the Center for Disease Control (CDC).$\textsuperscript{\ref{comm:us-data}}$ It contains the count of patient visits for ILI (positive cases) for each week and each state in the United States from 2010 to 2017. After removing a state with missing data we keep 49 states remaining in this dataset.
\item \textbf{Canada-Covid.} This dataset is publicly available at  JHU-CSSE.\footnote{https://github.com/CSSEGISandData/COVID-19} We collect daily COVID-19 cases from January 25, 2020 to January 10, 2022 in Canada (including 10 provinces and 3 territories). 
\end{itemize}

We adopt two metrics for evaluation that are widely used in epidemic forecasting, including the \textbf{Root Mean Squared Error (RMSE)} and the \textbf{Pearson's Correlation (PCC)}. RMSE measures the difference between predicted and true value after projecting the normalized values into the real range. PCC is a measure of the linear dependence between time series. For RMSE lower value is better, while for PCC higher value is better.




\subsection{Methods for Comparison}
We compared the proposed model with the following methods.

\begin{itemize}[leftmargin=*]
    \item \textbf{AR} The most classic statistical methods in time series analysis.
    \item \textbf{LRidge} The vector autoregression (VAR) with L2-regularization.
    \item \textbf{LSTNet ~\cite{lai2018modeling}} A deep learning model that combines CNN and RNN to extract short- and long-term patterns.
    \item \textbf{TPA-LSTM ~\cite{shih2019temporal}} An attention based LSTM network that employs CNN for pattern representations;
    \item \textbf{CNNRNN-Res ~\cite{wu2018deep}} A deep learning model that combines CNN, RNN, and residual links for epidemiological prediction.
    \item \textbf{SAIFlu-Net ~\cite{jung2021self}} A self-attention based deep learning model for regional influenza prediction.
    \item \textbf{Cola-GNN ~\cite{deng2020cola}} A deep learning model that combines CNN, RNN and GNN for epidemiological forecasting.
\end{itemize}

\subsection{Experimental Details}
All programs are implemented using Python 3.8.5 and PyTorch 1.9.1 with CUDA 11.1 (1.9.1 cu111) in an Ubuntu server with an Nvidia Tesla K80 GPU. Our source codes are publicly available.\footnote{https://github.com/Xiefeng69/SEFNet}

\textbf{Experimental setting}. All datasets have been split into training set(50\%), validation set(20\%) and test set(30\%). The batch size is set to 128. We use Min-Max normalization to convert data to [0,1] scale and after prediction, we denormalize the prediction value and use it for evaluation. The input window size $T$ is set to 20, and the horizon $h$ is set to \{3,5,10\} in turn. All the parameters of models are trained using the Adam optimizer with weight decay 5e-4, and the dropout rate is set to 0.2. We performed early stopping according to the loss on the validation set to avoid overfitting. The learning rate is chosen from \{0.01,0.005,0.001\}.

\textbf{Hyperparameters setting}. The hidden dimension of LSTM $D$ and attention layer $A$ is chosen from \{16,32,64\}. The number of LSTM layers $L$ is chosen from \{1,2\}. The number of kernels $K$ is chosen from \{4,8,12,16\}. The output dimension of Adaptive Max Pooling $P$ is chosen from \{1,3,5\}. The look-back window of AR component $q$ is chosen from \{0,10,20\}.

\subsection{Main Results}
We evaluate our model in short-term (horizon = 3) and long-term (horizon = 5,10) settings. \autoref{tab:result} summarizes the results of all methods. The large difference in RMSE values across different datasets is due to the scale and variance of the datasets, i.e., the scale of the Japan-Prefectures and Canada-Covid is greater than the US-Regions and US-States datasets, which is closely related to the prevalence of epidemics and population density. There is an overall trend that the prediction accuracy drops as the prediction horizon increases because the larger the horizon, the harder the problem.

We observe that the proposed SEFNet achieves the state-of-the-art results on most of the tasks. Traditional statistics methods (AR and LRidge) do not perform well in influenza-related datasets. The main reason is that they are based on oversimplified assumptions and only rely on historical records, they cannot model the strong seasonal effects in influenza datasets. For deep learning-based methods, the performance is improved since they make efforts to deal with nonlinear characteristics and complex patterns behind time series, However, some deep learning-based models work well on some datasets, while not well on others. 

\begin{enumerate}[leftmargin=*]
    \item \textbf{Performance.} The methods mainly focused on dependencies between regions/time series (Cola-GNN and TPA-LSTM) have better performance than the method mainly focused on dependencies between time points (LSTNet), which can point out that inter-series dependencies are quite valuable information. Our proposed model takes inter-series correlations between different regions and temporal relationships in a single region into consideration and carefully fuses them for prediction to achieve better performance.
    \item \textbf{Stability.} For Japan-Prefectures and Canada-Covid datasets, the prediction performance of many compared methods greatly decreases when the horizon increases. Because as the variance of the dataset increases, the fluctuation within time series and dependencies between regions are more intricate. SEFNet can make full use of dependencies information, so its predict error rises smoothly and slowly within the growing prediction horizon. Therefore, SEFNet has a better stability.
\end{enumerate}

Through this experiment, it can be concluded that SEFNet has better performance and stability in epidemic forecasting, especially in long-term prediction.

\subsection{Ablation Study}
In order to clearly verify that the above improvement comes from each added component, we conduct an ablation study on the US-Regions and US-States datasets. Specifically, we remove each component one at a time in SEFNet. We name the model without different components as follows:

\begin{itemize}[leftmargin=*]
    \item \textbf{w/oInter} The model without Inter-Series Embedding module.
    \item \textbf{w/oIntra} The model without Intra-Series Embedding module.
    \item \textbf{w/oAR} The model without the AR component.
    \item \textbf{w/oRAConv} The model uses 1$\times$3 convolution blocks only instead of Region-Aware Convolution.
    \item \textbf{w/oFusion} The model concatenates Inter-Series Embedding and Intra-Series Embedding directly instead of using Parametric Matrix Fusion.
\end{itemize}

\begin{figure}
  \centering
  \includegraphics[width=\linewidth]{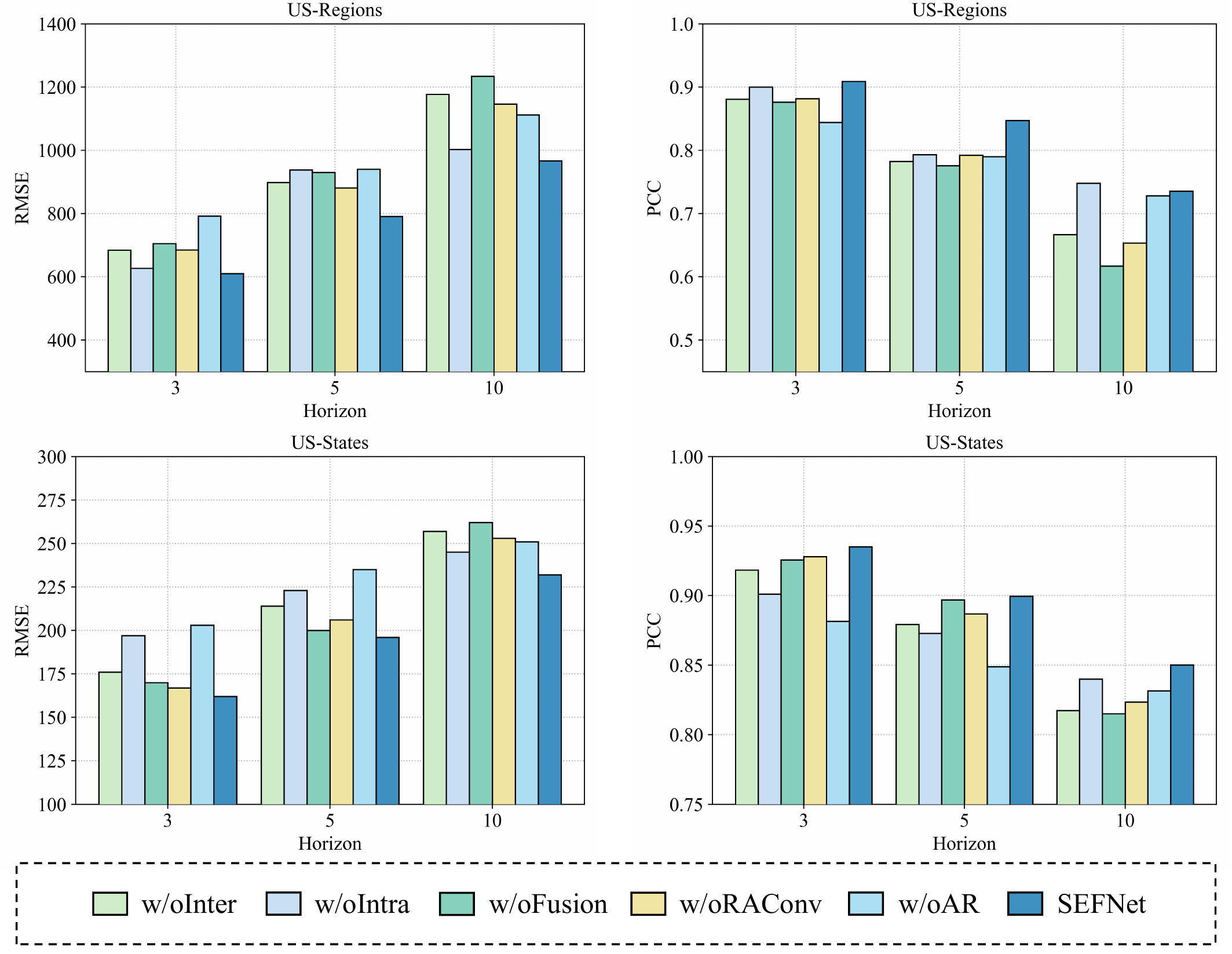}
  \caption{Results of the ablation studies on the US-Region (top) and US-States (bottom) datasets.}
  \label{fig:ablation}
\end{figure}


The ablation results are shown in \autoref{fig:ablation}. We highlight several observations from these results:
\begin{enumerate}[leftmargin=*]
    \item The full SEFNet model achieves almost the best results. 
    \item Directly concatenating two feature embeddings will bring performance drops, while the fusion method used in SEFNet can bring performance gains, especially in long-term prediction (horizon=5,10). Because when the horizon increases, the difficulty of prediction will increase accordingly, and the complex dependencies among regions are more difficult to detect. Parametric-matrix fusion adaptively learns the importance of inter- and intra-series embeddings, which facilitates capturing complex and potential relationships in long-term prediction.
    \item Compared with single-scale convolutions, Region-Aware Convolution brings performance improvement by capturing and aggregating multi-scale features, which proves its strong feature representation power.
    \item Removing the AR component from the full model caused significant performance drops, showing the crucial role of the AR component in general.
\end{enumerate}

This ablation study concludes that our model design is the most robust across all baselines, especially with large horizons.

\section{Conclusion}
In this paper, we propose an Inter- and Intra-\underline{\textbf{S}}eries \underline{\textbf{E}}mbeddings \underline{\textbf{F}}usion \underline{\textbf{Net}}work (SEFNet) for epidemic forecasting. We first extract inter- and intra-series embeddings from two parallel modules. Specifically, in inter-series embedding module, we design a Region-Aware Convolution component that is better to extract feature representations of time series and capture the dynamic dependencies among regions. Then, we fuse two embeddings through parametric-matrix fusion for prediction. To further enhance the robustness, we apply an AutoRegressive component as the linear part. Experiments on four real-world epidemic-related datasets show the proposed model outperforms the state-of-the-art baselines in terms of performance and stability. In future work, we plan to delve into the dynamic dependencies and mutual influences among regions.

\begin{acks}
This work is supported by the Key R\&D Program of Guangdong Province No.2019B010136003 and the National Natural Science Foundation of China No. 62172428, 61732004, 61732022.
\end{acks}


\bibliographystyle{ACM-Reference-Format}

\end{document}